# Simple Full-Spectrum Correlated *k*-Distribution Model based on Multilayer Perceptron

Xin Wang[a], Yucheng Kuang[b], Chaojun Wang[c]*, Hongyuan Di[c], Boshu He[c]

[a]School of Information and Communication Engineering, Key Laboratory of Modern Measurement & Control Technology, Ministry of Education, Beijing Information Science and Technology University, Beijing 102206, People's Republic of China

[b]Beijing Aerospace Petrochemical EC and EP Technology Corporation Limited (BAEEC), Beijing 100176, People's Republic of China

[c]Institute of Combustion and Thermal Systems, School of Mechanical, Electronic and Control Engineering, Beijing Jiaotong University, Beijing 100044, People's Republic of China

**Abstract:** While neural networks have been successfully applied to the full-spectrum *k*-distribution (FSCK) method at a large range of thermodynamics with *k*-values predicted by a trained multilayer perceptron (MLP) model, the required *a*-values still need to be calculated on-the-fly, which theoretically degrades the FSCK method and may lead to errors. On the other hand, too complicated structure of the current MLP model inevitably slows down the calculation efficiency. Therefore, to compensate among accuracy, efficiency and storage, the simple MLP designed based on the nature of FSCK method are developed, i.e., the simple FSCK MLP (SFM) model, from which those correlated *k*-values and corresponding *ka*-values can be efficiently obtained. Several test cases have been carried out to compare the developed SFM model and other FSCK tools including look-up tables and traditional FSCK MLP (TFM) model. Results show that the SFM model can achieve excellent accuracy that is even better than look-up tables at a tiny computational cost that is far less than that of TFM model. Considering accuracy, efficiency and portability, the SFM model is not only an excellent tool for the prediction of spectral properties, but also provides a method to reduce the errors due to nonlinear effects.

---

* Corresponding author. Email: cjwang@bjtu.edu.cn. Tel.: +86-10-5168-8542. Fax: +86-10-5168-8404.

# 1. Introduction

Radiation, as the dominant mode of heat transfer in high-temperature environments, plays an important role in a variety of energy conversion technologies, including combustion, gasification, solar thermochemical conversions, etc [1-5]. Accurate radiation modelling is not only important for those energy devices to shorten the design cycle but also can provide more insight beyond the capability of experiments. However, due to the complex nature, it is difficult to model radiation, of which the major challenge is the radiative properties for gases, e.g., $CO_2$, $H_2O$ and CO. Because of oscillating variation across the spectrum, their radiative properties display very strong spectral or 'nongray' behavior.

The most accurate spectral properties can be predicted by the so-called line-by-line (LBL) method [6] at a high spectral resolution but with a huge computational cost in both time and memory. To model nongray behaviors efficiently while keep accuracy, the full-spectrum $k$-distribution (FSK) method was proposed by Modest [7]. By employing the reordering, the absorption coefficients can be transformed into the monotonic $k$-distribution across an artificial spectrum, on which the integration can be performed. This can reduce the evaluations of radiative transfer equation (RTE) from millions to around ten without losing accuracy. Furthermore, by employing correlation principle, Wang et al. [8-10] constructed the full-spectrum look-up table (referred to as table in the following) to avoid the time-consuming processes of both mixing and assembling $k$-distributions, achieving almost LBL accuracy with only a tiny fraction of the LBL computational cost.

While the table is both accurate and efficient, the huge size that is more than 20 GB for all thermodynamic states hindered their applications to engineering use. On the other hand, the gap between radiation researchers and engineers who use the radiation models further makes the table inconvenient. To mitigate this issue, the machine learning method was applied to the FSCK method and a multi-layer perceptron (MLP) model based on neural networks occupied only around 40 MB was developed by [11, 12], namely traditional FSCK MLP (TFM) model in this work, which can predict as many as 32 correlated $k$-values at each thermodynamic state in high accuracy compared to those from the table. The required nongray stretching factors ($a$-values) then need to be calculated on-the-fly based on 32 correlated $k$-values. While 32 values can achieve high accuracy for the prediction of $a$-values, the on-the-fly calculation theoretically deviates the definition of

*a*-value that is a ratio between two differentials of cumulative *k*-distributions. With limited *k*-values, the differential can be only replaced by the difference, which degrades the FSCK method into a lower order form [13]. On the other hand, it is impossible to use all 32 correlated *k*-values during radiative calculations due to both CPU and memory considerations. However, due to accuracy requirement of *a*-values, all 32 values need to be obtained when using the TFP model, which is inevitably time-consuming.

Another attempt of neural network applications to spectral model was made by Sun et al. [14]. They used the neural network method to calculate inverse absorption line blackbody distribution function (ALBDF) so that the cross-sections can be obtained according to a specified ALBDF. Then, the SLW model based on neural network was developed. However, only cross-sections of single gas (either $CO_2$ or $H_2O$) can be obtained and the mixing scheme must be used for mixtures, which both increases CPU cost and may lead to errors. While the efficiency was reported to be significantly reduced, the CPU time required by the neural network based SLW model still stay in a large level compared to that of the FSCK table.

It is no doubt that the neural network is indeed a good tool to achieve both accurate and efficient spectral prediction, and most importantly, can alleviate data storage to a large extent. The key is how to compensate among accuracy, efficiency and storage, prompted by which a simple FSCK MLP (SFM) model available for a large range of thermodynamic states is developed in this work. Both correlated *k*-values and *ka*-values can be accurately obtained from the SFM model at a tiny computational cost. With the storage less than 0.5 MB, the SFM is not only easy to implement to other platforms, but also provide a way to alleviate the errors due to nonlinear effects.

## 2. Full-spectrum correlated *k*-distribution method

Full-spectrum correlated *k*-distribution (FSCK) method can be regarded as one of nonhomogeneous extensions of full-spectrum *k*-distribution (FSK) method by using correlation principle. To handle complex spectral behaviors, the FSCK method first reorders the absorption coefficients by constructing a Planck-function-weighted *k*-distribution defined as [15]

$$f_{T,\boldsymbol{\phi}}(k) = \frac{1}{I_b(T)} \int_0^\infty I_{b\eta}(T) \delta(k - \kappa_\eta(\boldsymbol{\phi})) \mathrm{d}\eta \tag{1}$$

where *T* represent the Planck function temperature; $\boldsymbol{\phi}$ represents the local thermodynamic state vector including pressure, temperature and species concentration; *k* represents a nominal value between the minimum and maximum absorption coefficients $\kappa_\eta$ at state $\boldsymbol{\phi}$; $I_b$ represents the blackbody intensity;

$\eta$ and $\delta(\cdot)$ are the wavenumber and the Dirac-Delta function, respectively.

Then, the cumulative *k*-distribution can be defined as

$$g_{T,\phi}(k) = \int_0^k f_{T,\phi}(k')\mathrm{d}k' \tag{2}$$

Thus, $g$ represents a fraction of spectrum whose absorption coefficients lie below a specific $k$ and, consequently, $0 \le g \le 1$. By inverting Eq. (2), a smooth, monotonically increasing function, $k_{T,\phi}(g)$, can be obtained, with minimum and maximum values identical to those of the absorption coefficients.

With the cumulative *k*-distribution *g*, the radiative transfer equation (RTE) in an absorbing and emitting medium can be transformed into a reference $g_0$-space using the correlation principle [15], i.e.,

$$\frac{\mathrm{d}I_g}{\mathrm{d}s} = k^*(\phi)\left[a(T,T_0,g_0)I_b(T) - I_g\right] \tag{3}$$

where

$$I_g = \left.\int_0^\infty I_\eta \delta\left(k - \kappa_\eta(\eta,\phi_0)\right)\mathrm{d}\eta \middle/ f_{T_0,\phi_0}(k) \right. \tag{4}$$

$$a(T,T_0,g) = \frac{\mathrm{d}g_{T,\phi_0}(k)}{\mathrm{d}g_{T_0,\phi_0}(k)} = \frac{\mathrm{d}g_{T,\phi}(k^*)}{\mathrm{d}g_{T_0,\phi}(k^*)} \tag{5}$$

$k^*$ is a correlated *k*-value defined in [13], $T_0$ is the reference temperature and *s* is the path length.

For the FSCK method, finding the appropriate *k\** and *a* is the key point to achieve high accuracy, for which a large number of implementations has been proposed over the past twenty years. The authors [13] compared them in detail and concluded that one important criterion is the emission should be preserved, which can be achieved by the correlation principle even for the practical absorption coefficients that are not perfectly correlated. This leads to several accurate implementations, i.e., Methods 2~4 mentioned in [13]. Technically, each of Methods 2~4 is able to provide *k\** and *a* at a large range of thermodynamic states used for training the machine learning model. The major difference between them is the required number of evaluations, of which Methods 3 and 4 are less than those of Method 2. Less evaluations mean the calculations could be more efficient. On the other hand, considering Method 3 currently is available for not only the FSCK method but also the spectral line based weighted sum of gray gases (SLW) method, the FSCK-3 implementation based on Method 3 is selected in this work for the development of machine learning model. This extends the applicability of the developed model since the SLW method as a mathematically equivalent method to the FSCK method can also share the training data generated based on Method 3.

# 3. Simple FSCK MLP model (SFM)
## 3.1. Structure

Similar to the TFM model, the SFM model is still based on the MLP model with a number of neurons. For each neuron, the input is the sum of weighted output (i.e., $w_i x_i$) derived from neurons in the previous layer as shown in Fig. 1; the output is obtained by adding the bias $b$ (that is commonly the same for the neurons in the same layer) and then transformed by the so-called activation function $f$ (that is the same either for the neurons in the same layer or for all neurons); finally, the output of this neuron will serve as one of input for neurons in the next layer. Therefore, the storage of each neuron mainly concerns the values of weight and bias.

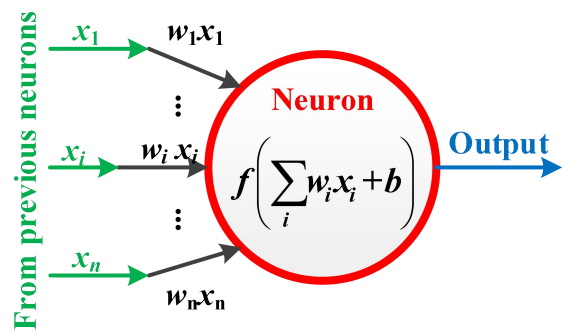

Fig. 1 The diagram of calculations for one neuron in an artificial neural network

Those neurons in the MLP model are then categorized into one input layer, one output layer and several hidden layers, each of which is fully connected to the adjacent layer. In this manner, a direct and acyclic neuron network can be constructed, an example of which can be found in Fig. 2. Commonly for an MLP model, the more the number of neurons is, the more accurate results and storage the MLP model give but the less efficiency it achieves. Therefore, it is important to design a smart structure for compensating among accuracy, efficiency and storage.

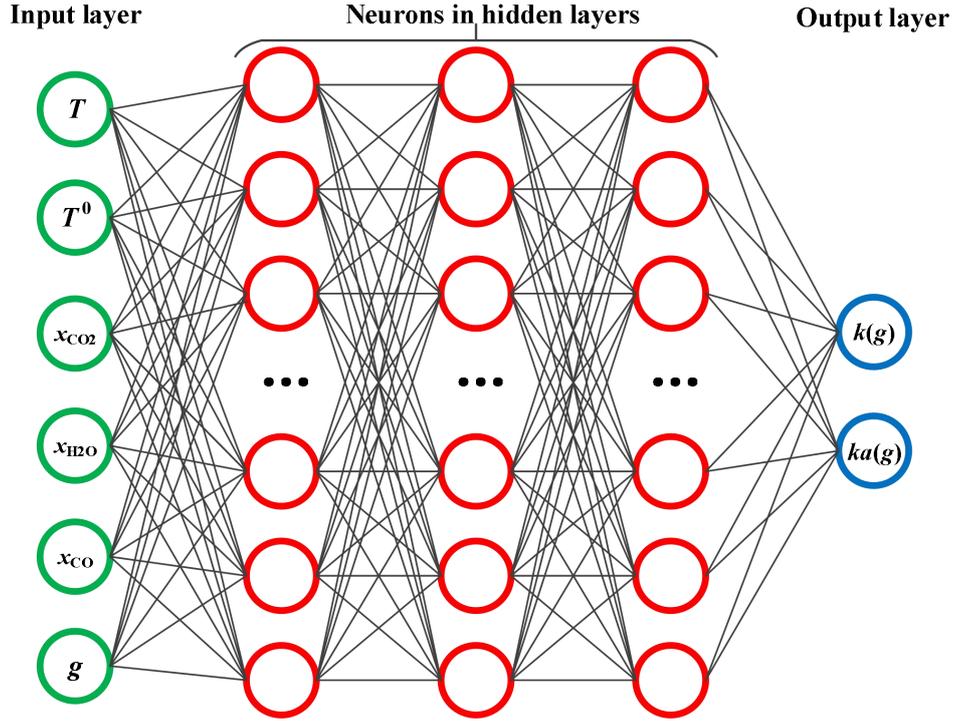

Fig. 2 The structure of SFM model based on the MLP neural network architecture

In the TFM model, the input layer only contains the thermodynamic state, i.e., pressure, temperature and concentrations while the output is a $k$-distribution at the corresponding state consisted of several discreate $k$-values. Since the required $a$-values are calculated from those $k$-values, the number of neurons in the output layer of TFM model is as many as 32 to keep the accuracy of $a$-values. This inevitably increases the complexities of TFM model. To guarantee the accuracy, 5 or 6 hidden layers are chosen with more than 200 neurons in each layer. While the storage is reduced to approximate 43 MB, the efficiency becomes unacceptable. In addition, it is unnecessary to obtain 32 values and commonly 8 values are enough during practical radiative calculations; however, due to the requirement of $a$-values, the TFM model has to load all 32 values and then obtain 8 values by interpolations, which also leads to additional computational cost.

To our best knowledge, there is currently no theories for how to set the number of hidden layers and neurons in each layer. The only way is trial-and-error. For the SFM model considered in this work, the number of hidden layers is set to 3, which is a common value from experiences. Then, how to design the input and output layers is the key for the SFM model.

As shown in Fig. 2, the input layer of SFM model also include five parameters of thermodynamic state: Planck function temperature $T$, reference temperature $T_0$, mole fractions of $CO_2$, $H_2O$ and CO. The pressure is not included in the input layer in order to decrease the number of input neurons. On the other hand, the constant pressure is found in most practical applications, so the structure of the SFM model shown in Fig. 2 can be trained at each pressure. Furthermore, the cumulative $k$-distribution $g$ becomes another neuron in the input layer, the reason of which is to significantly reduce the number of neurons in the output layer. In this way, the SFM model can directly predict a single $k$-value at certain $g$ rather than a $k$-distribution at a state.

Compared to the $k$-distribution (or value) that is monotonically increasing with $g$ across a large range of magnitudes, the $a$-value in FSCK method shows an extremely oscillating trend with a relatively small range of magnitudes as shown in Fig. 3a. This brings in a fact that the $a$-value is more difficult to be both trained and predicted by the neuron networks than the $k$-value. Considering the $a$-value is always used along the $k$-value in the emission term as shown in Eq. (3), it is more appropriate to combine them, i.e., train and predict the $ka$-value rather than the $a$-value only. Figure 3b gives the distribution of $ka$-value at a certain state. Although the strict monotonicity is still not satisfied, the trend of $ka$-value across $g$ becomes more regular than that of $a$-value. Moreover, comparisons of $k$-value in Fig. 3a and $ka$-value in Fig. 3b show their weak similarity, which means at a certain $g$, the gap between $k$ and $ka$-values is relatively small. This can be explained by the small range of magnitudes for $a$-values across most of $g$ (although oscillating). As a result, the output layer of SFM model has two neurons as shown in Fig. 2: one is $k$ and the other one is $ka$. This not only avoids directly predicting the oscillating $a$-values, but also employs the weak similarity to use only one neural network for the prediction of both $k$ and $ka$ rather than two neural networks for those separate predictions of $k$ and $ka$. With these designs mentioned above, the storage of the SFM model at 1 atm is only 0.43 MB.

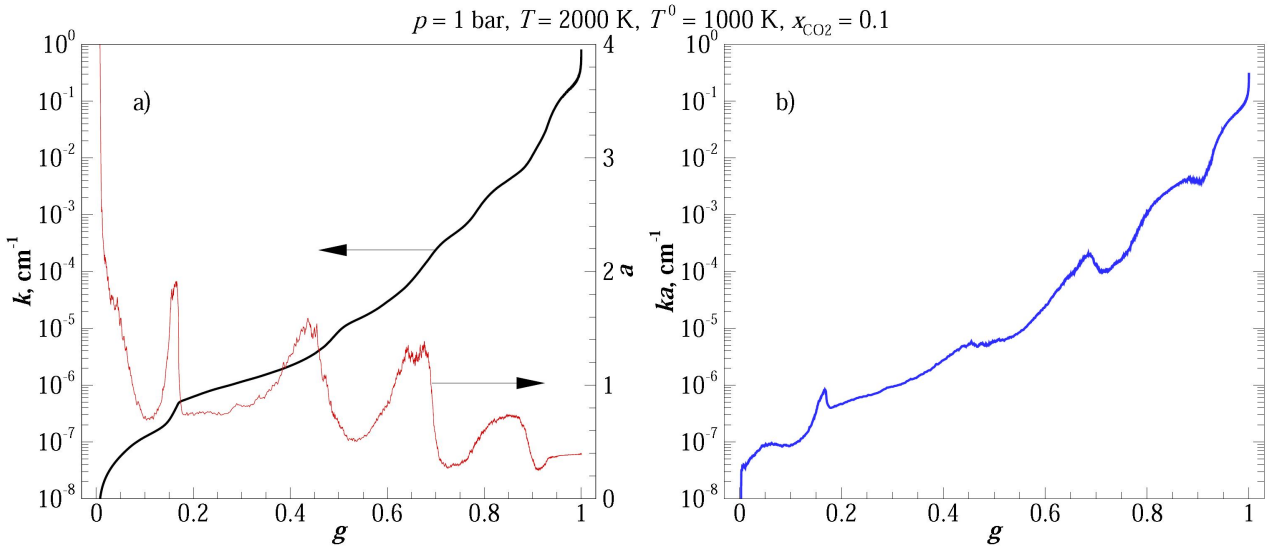

Fig. 3 a) $k$-distribution and corresponding nongray stretching factor $a$-values as well as b) $ka$-distribution. Local and reference state are 10% $CO_2$ in $N_2$ at 2,000 K and 1,000 K; $p = 1$ bar

## 3.2. Training

The training data includes the thermodynamic states and corresponding $k$-values as well as $ka$-values. In order to make available for most practical applications at 1 atm, a large number of thermodynamic states with temperature ranging from 300 K to 3000 K and mole fractions ranging from 0 to 1 (for CO, the range is from 0 to 0.5 since higher values can be barely found in combustion system) are considered here. The thermodynamic states are then selected randomly from the data values listed in Table 1. The reason for a number of small mole fractions is to decrease these nonlinear effects caused by gases [8]. The number of mole fractions is approximate double of that used to construct both table [8] and TFM model [11]. In this work, 50,000 random thermodynamic states are generated and for each state both $k$- and $ka$-values corresponding to 8-point Gauss-Chebyshev quadrature are generated using the LBL database updated to HITEMP2010 [16]. Since the quadrature $g$ is also an input neuron, this results in 400,000 thermodynamic states as input and 400,000 $k$-values (and $ka$-values) as output during the training process.

Table 1 Thermodynamic states used for generation of training data

| Parameters | Range | Values | Number of points |
|---|---|---|---|
| Species | $CO_2$, $H_2O$ and CO | - | 3 |
| Planck temperature | 300~3,000 K | Every 100 K | 28 |
| Reference temperature | 300~3,000 K | Every 100 K | 28 |
| Mole fraction of $CO_2$ | 0.0~0.05 | Every 0.005 | 18 |
|  | 0.1~0.2 | Every 0.05 |  |
|  | 0.25~1.0 | Every 0.25 |  |
| Mole fraction of $H_2O$ | 0.0~0.05 | Every 0.005 | 18 |
|  | 0.1~0.2 | Every 0.05 |  |
|  | 0.25~1.0 | Every 0.25 |  |
| Mole fraction of CO | 0.0~0.05 | Every 0.01 | 9 |
|  | 0.1~0.5 | [0.1, 0.25, 0.5] |  |

The training process in this work is done by Keras [17], which is a convenient deep learning tool and provide amounts of method interface for machine learning. To significantly improve the efficiency while keep the accuracy, the number of neurons in each hidden layer of the SFM model is limited to 120, which is obtained by trial-and-error. The activation function for each layer is the rectified linear unit (Relu) and the backpropagation algorithm in conjunction with the stochastic gradient-based optimizer (Adam) [18] is used to adjust appropriate weights and biases of neurons. The Bayesian optimization process proposed in [19] was employed to choose the optimal SFM neural network hyperparameters, i.e., the initial learning rate and regularization factor, the latter of which is applied to each hidden layer for the sake of avoiding overfitting. In this work, the Bayesian optimization was applied to minimize a custom $R^2$-score based training metric, which is defined as,

$$\text{Metric} = \frac{\sum_{i=1}^{n}(y_i - \hat{y}_i)^2}{\sum_{i=1}^{n}(y_i - \bar{y})^2} \tag{6}$$

where $\hat{y}_i$ is the value predicted by the training model; $y_i$ is the corresponding true value and $\bar{y}$ is defined as

$$\bar{y} = \frac{1}{n}\sum_{i=1}^{n} y_i \tag{7}$$

The smaller the Metric is, the better prediction the SFM model gives. The training stops when 100 samples of Bayesian optimization are finished. The optimal hyperparameters of SFM model are $1.246 \times 10^{-3}$ for initial rate and $10^{-7}$ for regularization factor with the Metric lower than $5 \times 10^{-4}$.

**3.3. Implementation**

All the training processes described above are carried out using Python, which is an interpreted programming language and does not need to be compiled. Considering the radiative module is currently written in Fortran, it is straightforward to construct a Python-Fortran interface for transferring data forth and back during radiative calculations. However, the interpreted nature of Python inevitably increases computational cost when employing Python-Fortran interface. In addition, the portability of Python-Fortran interface to other platforms of computational fluid dynamics (CFD) is somewhat complex and inconvenient because those Python packages always need to be pre-installed before running the model. To end this, the machine learning model trained by Python is programmed into several Fortran modules including initialization, calculation and finalization, which forms the SFM model. The weights and biases for all neurons are stored in several data files. During radiative calculations, these data files are firstly written into memory by running initialization module, followed by the calculations in each neuron. Finalization module is called when radiative calculations are completed. Currently, the SFM model has been compiled into a stand-alone library and implemented to the CFD platform OpenFOAM [20] to test the performance for predicting nongray properties during radiation modeling. It should be noted that with the same logic, the model can also be implemented to other CFD platforms.

**4. Results and discussions**

**4.1. Verifications *k*- and *ka*-distribution**

Comparisons of *k*- and *ka*-distributions calculated from either exact FSCK model and SFM model at a certain thermodynamic state are shown in Fig. 4. The state is chosen in the manner that the state values are not in Table 1 to guarantee the SFM model never sees this state. The exact FSCK model directly reorders absorption coefficients at spectral level based on the LBL database updated to HITEMP2010. Therefore, those *k*- and *ka*-distributions calculated by the exact FSCK model give the benchmark calculations. It can be observed in Fig. 4 that most of those discrete *k*- and *ka*-values predicted by the SFM model overlap very well with the benchmark results and a few of them may give some errors. This is mainly because the predictive ability of SFM model is somewhat affected by the structure that requires to sacrifice part of accuracy to keep efficiency. However, those errors stay in a low level and do not give much influence during radiative calculations. This is demonstrated in Fig. 5 that gives the emission, radiative heat flux *q* and negative radiative heat source $\nabla \cdot q$ of the gas mixture across a 1D slab using the *k*- and *ka*-values in Fig. 4. For comparisons, the exact FSCK method also employs the 8-point Gauss-Chebyshev quadrature. The RTE for both models is solved by the exact solution. It is clear that the results using the SFM model show good agreement with those using the exact FSCK method, with the maximum relative error less than 2% across the slab. Similar phenomenon can be also seen in other radiative calculations using the SFM model, which illustrates the SFM model is able to predict *k*- and *ka*-values in acceptable accuracy.

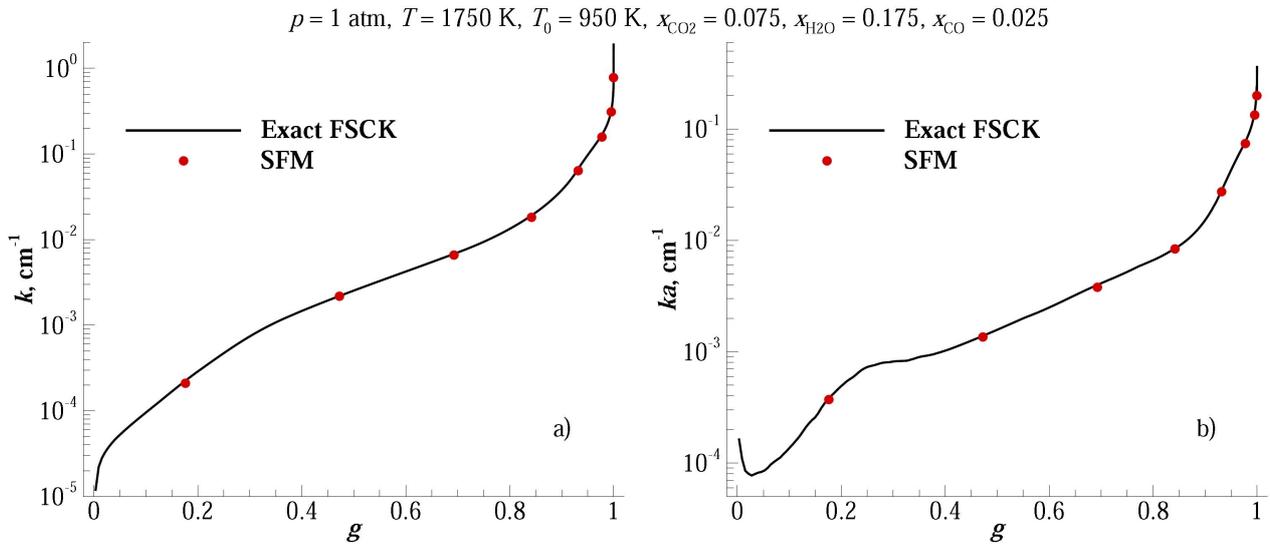

Fig. 4 Comparisons of a) *k*-distribution and b) *ka*-distribution calculated by exact FSCK method and SFM model. Local and reference state are 7.5% $CO_2$-17.5% $H_2O$-2.5% CO in $N_2$ at 1,750 K and 950 K; $p = 1$ atm

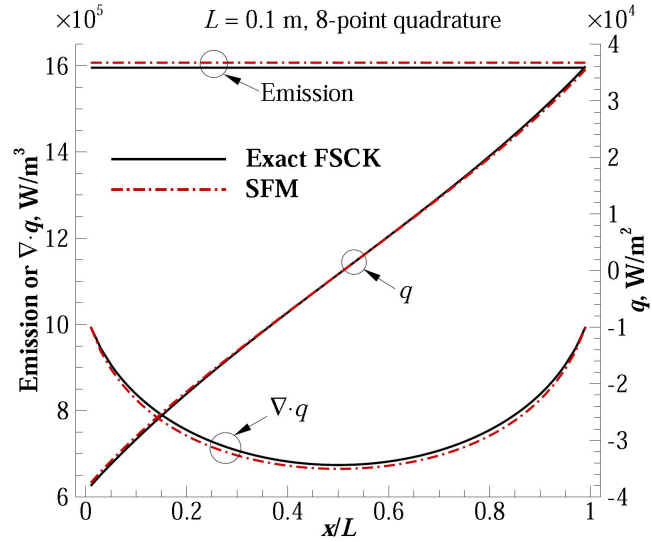

Fig. 5 Emission, radiative heat flux and negative radiative heat source for a 1D slab using exact FSCK method and SFM model at the state shown in Fig. 4

Since *ka*-values do not show the strict monotonic trend, they are more difficult to be trained than *k*-values. To further verify those *ka*-values predicted by the SFM model, the Planck mean absorption coefficients $\kappa_P$ using different models for a gas mixture with different Planck function temperatures at a fixed reference state are compared in Fig. 6. Except for the benchmark LBL results, all FSCK calculations (table, TFM and SFM) employs an 8-point Gauss-Chebyshev quadrature. At high Planck function temperatures ($T > 1,000$ K), discrepancies among the $\kappa_P$ calculated by different methods are barely noticeable. When the Planck function temperatures are low, the obvious errors can be seen for all FSCK calculations. The common reason is due to the quadrature errors. For both TFM and SFM models, they show different performances: except for 300 K at which the emission can generally be neglected, the SFM model give better predictions of $\kappa_P$. This implies that the SFM model is not only more robust than the TFM model but also sufficient for radiative calculations.

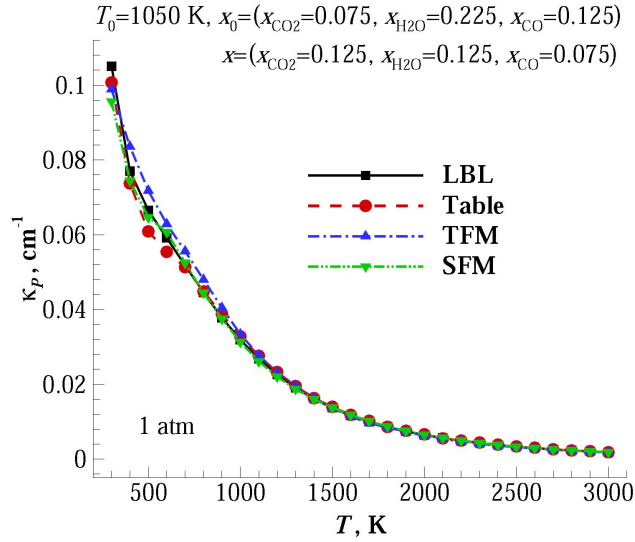

Fig. 6 Planck mean absorption coefficients calculated by LBL, table, TFM model and SFM model for gas mixtures with different Planck function temperatures at a fixed reference state; $p = 1$ atm

To test the efficiency of the SFM model, CPU times of generating $k$-distributions for a large number of gas mixtures at random thermodynamic states are compared by using the table, TFM model and present SFM model. The pressure for all $k$-distributions is fixed to 1 atm and the temperature is arbitrarily ranging from 300 K~3,000 K. The mole fractions of $CO_2$ and $H_2O$ are arbitrarily ranging from 0 to 1 while those of CO are arbitrarily ranging from 0 to 0.5. Table 2 gives comparisons of CPU times for generating 10,000 arbitrary $k$-distributions and corresponding $a$-values based on three different methods. The required CPU time for the table is the shortest since only linear interpolations are required. Using the TFM model considerably increases the CPU time because it contains more than 5 hidden layers with a large number of neurons. Such a large structure makes each generation of $k$-distribution and corresponding $a$-values time-consuming. By contrast, CPU time of generation using the SFM model is much smaller than that using the TFM model, which mainly due to the simple structure. While it still costs more computational time than the table, the SFM model achieves the huge reduction of CPU time, which is definitely available to the engineering applications.

Table 2 CPU times of generating *k*-distributions at 10,000 arbitrary thermodynamic states using different FSCK models

| Models | CPU, s |
|--------|--------|
| Table  | 0.09   |
| TFM    | 62.18  |
| SFM    | 0.59   |

**4.2. Flame calculations**

In order to test the performance of SFM model in realistic applications, radiation from a jet flame is considered in this section. The jet flame is derived from Sandia Flame D [21] by artificially increasing the geometry by four times, leading to strong radiation effects that can be found in practical combustion configurations. The simulations of the flame are carried out on a two-dimensional axisymmetric mesh. Only the quasi-steady time-averaged fields of temperature and species mass fraction shown in [8] are used in this work to compare radiative calculations obtained by different models. In addition, turbulence-radiation interaction is not considered here. Emission and $\nabla \cdot q$ are calculated by different models with RTE solved by the P1 method and no feedback to the flame. The total pressure is 1 atm and three gases i.e., $CO_2$, $H_2O$ and $CO$, are considered in the calculations. All FSCK calculations including table, TFM and SFM models employ an 8-point Gauss-Chebyshev quadrature for spectral integrations and are compared with benchmark LBL calculations at different locations (three axial and one radial locations).

The results using both table and TFM model give some discrepancies compared to the LBL results as shown in Fig. 7, one main reason of which comes from the nonlinear effects of *k*-distribution resulted from limited state values in the table. For the TFM model, the nongray stretching factors, i.e., *a*-values, are calculated on-the-fly by using 32 discrete *k-g* values, which make the differential calculations shown in Eq. (5) become difference calculations; this somewhat degrade the FSCK method and may lead to errors. In addition, training data of the TFM model is constructed using exactly the same state values in the table; however, the nature of regression must cause another error source. These lead to a fact that the TFM model shows a slightly inferior performance compared to the table for predictions of both emission and $\nabla \cdot q$ as shown in Figs. 7b and 7c. In contrast, the training data of the SFM model is constructed by using the extension of state values as shown in Table 1. Increasing the state values at low mole fractions is able to alleviate nonlinear effects to an extent. Also, the *a*-values predicted by the SFM model are directly obtained by those *ka*-values without degrading. While errors due to regression still exists, the SFM model gives the best performance among different FSCK models with almost all predictions overlapping well with the LBL results. This not only demonstrates the robust ability of the SFM model for the radiative calculations in real scenarios, but also provides a new way to reduce the nonlinear effects.

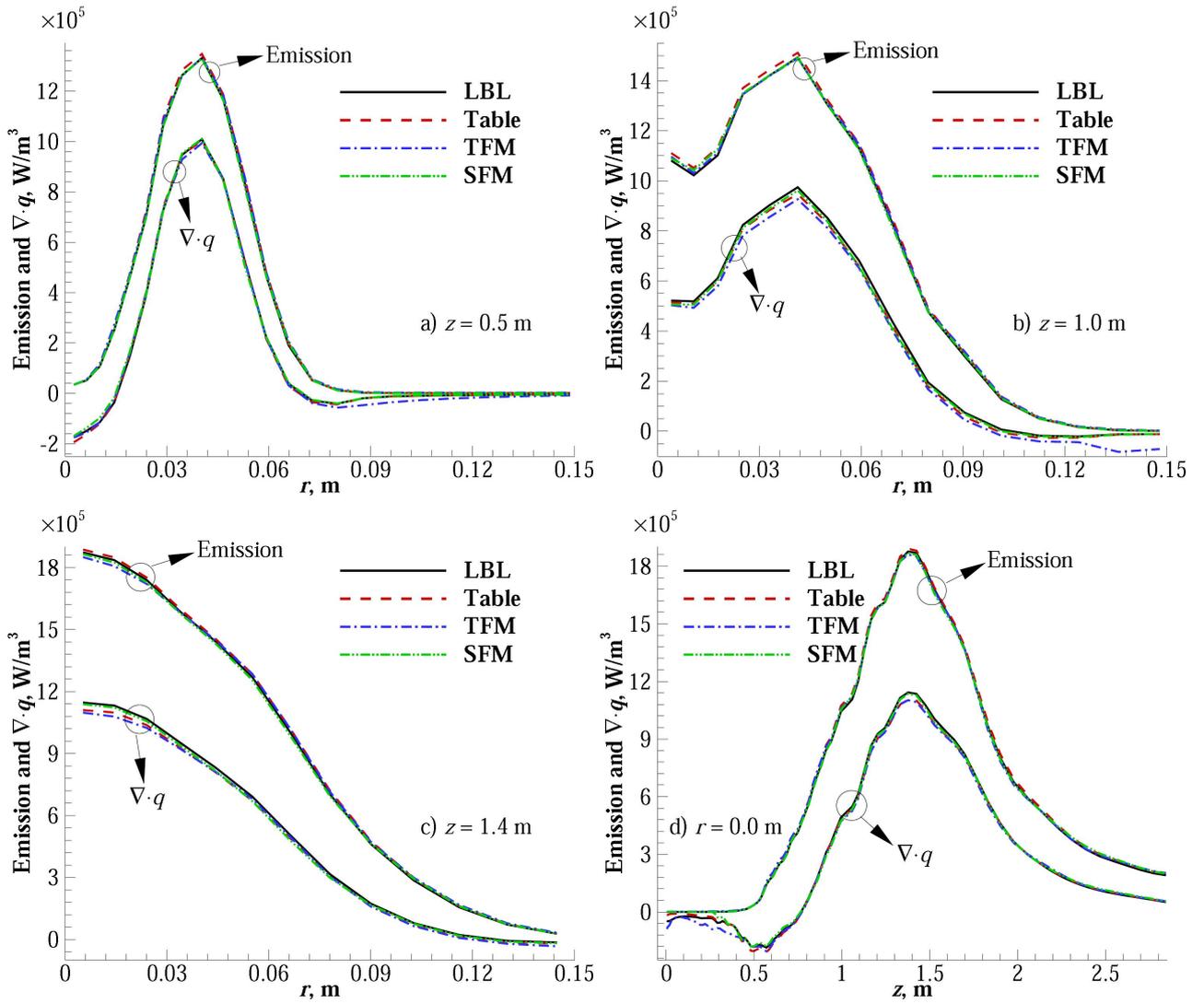

Fig. 7 Comparisons of emission and negative radiative heat source for Sandia flame D4 using different models at four locations, (a) $z = 0.5$ m, (b) $z = 6$ m, (c) $z = 1.4$ m and (d) $r = 0.0$ m.

CPU times for radiative calculations of Sandia flame D4 using different FSCK models are given in Table 3. One core of the 64-bit AMD Ryzen Threadripper PRO 3955WX CPU is employed in this work. As expected, due to complex structure with a large number of neurons, the TFM model requires the largest CPU time. When the spectral model becomes the SFM model, the CPU time reduces to only one tenth of that using the TFM model. Because of the inevitable sacrifice by compensating between accuracy and efficiency, CPU cost of SFM model is still larger than that of table. Considering the accuracy, efficiency, storage and portability, the current SFM model is suitable to be alternatives for modelling radiation.

Table 3 CPU comparisons of flame calculations using different FSCK models

| Models | CPU, s |
|--------|--------|
| Table  | 0.47   |
| TFM    | 23.07  |
| SFM    | 2.00   |

## 5. Conclusions

In this work, an accurate, efficient, compact as well as portable FSCK model based on the MLP neural network approach for gas mixtures, i.e., SFM model, is developed. Inspired by the nature of FSCK method, the structure of the SFM model is designed in the way that the input layer contains not only state values but also quadrature $g$ and the output layer has two neurons representing $k$- and $ka$-values, respectively. The SFM model is then trained using an extension of database with more data values putting in small mole fractions. Sample and flame calculations show the SFM model can be used to accurately predict both $k$- and $ka$-values for gas mixture. While the efficiency is still lower than that of table, the SFM model has improved a lot compared to the traditional version (TFM model). The size of SFM model is only about 0.43 MB that is significantly smaller than the TFM model. The SFM model consists of several stand-along Fortran modules, which can be easily compiled into portable library for the implementation to different CFD platforms. In addition, a new way by putting more low mole fractions for construction of training data is proved to be able to alleviate the nonlinear effects, which is expected to overcome the nonlinear effects found in most of FSCK table.

**Acknowledgments**

This work was supported by the National Natural Science Foundation of China (NSFC, 52006007), R&D Program of Beijing Municipal Education Commission (KM202311232012) and Open Fund of IPOC (BUPT).